\documentclass[10pt]{wlscirep}
\usepackage[utf8]{inputenc}
\usepackage{float}
\usepackage[T1]{fontenc}
\usepackage[draft]{changes}
\usepackage{enumitem}
\usepackage{comment}
\newcommand{\citep}{\cite} %

\title{
GPT detectors are biased against non-native English writers 
}

\author[1*]{Weixin Liang}
\author[1*]{Mert Yuksekgonul} %
\author[2*]{Yining Mao} %
\author[2*]{Eric Wu} %
\author[1,2,3,+]{James Zou} %

\affil[1]{Department of Computer Science, Stanford University, Stanford, CA, USA}
\affil[2]{Department of Electrical Engineering, Stanford University, Stanford, CA, USA}
\affil[3]{Department of Biomedical Data Science, Stanford University, Stanford, CA, USA}

\affil[+]{Correspondence should be addressed to: jamesz@stanford.edu}
\affil[*]{these authors contributed equally to this work}

\begin{abstract}

The rapid adoption of generative language models has brought about substantial advancements in digital communication, while simultaneously raising concerns regarding the potential misuse of AI-generated content. Although numerous detection methods have been proposed to differentiate between AI and human-generated content, the fairness and robustness of these detectors remain underexplored. In this study, we evaluate the performance of several widely-used GPT detectors using writing samples from native and non-native English writers. Our findings reveal that these detectors consistently misclassify non-native English writing samples as AI-generated, whereas native writing samples are accurately identified. Furthermore, we demonstrate that simple prompting strategies  can not only mitigate this bias but also effectively bypass GPT detectors, suggesting that GPT detectors may unintentionally penalize writers with constrained linguistic expressions. Our results call for a broader conversation about the ethical implications of deploying ChatGPT content detectors and caution against their use in evaluative or educational settings, particularly when they may inadvertently penalize or exclude non-native English speakers from the global discourse. 
The published version of this study can be accessed at: 
\url{www.cell.com/patterns/fulltext/S2666-3899(23)00130-7}

\end{abstract}
\begin{document}
\maketitle
\thispagestyle{empty}

\section*{Introduction}

Generative language models based on GPT, such as ChatGPT~\citep{ChatGPT}, have taken the world by storm. Within a mere two months of its launch, ChatGPT attracted over 100 million monthly active users, making it one of the fastest-growing consumer internet applications in history~\citep{reuters-chatgpt-sets-record-fastest-growing-user-base,forbes-chatgpt-hits-100-million}. While these powerful models offer immense potential for enhancing productivity and creativity~\citep{lee2022evaluating,chatgpt-pass-medical-exam,chatgpt-wharton-mba}, they also introduce the risk of AI-generated content being passed off as human-written, which may lead to potential harms, such as the spread of fake content and exam cheating~\citep{Nature-news-Abstracts-written-by-ChatGPT-fool-scientists,Abstracts-written-by-ChatGPT-fool-scientists,fake-news-66,nature-editorial-ChatGPT-tools,ICML2023LLMPolicy}.

Recent studies reveal the challenges humans face in detecting AI-generated content, emphasizing the urgent need for effective detection methods~\citep{Nature-news-Abstracts-written-by-ChatGPT-fool-scientists,Abstracts-written-by-ChatGPT-fool-scientists,fake-news-66,huaman-detect-gpt3}. Although several publicly available GPT detectors have been developed to mitigate the risks associated with AI-generated content, their effectiveness and reliability remain uncertain due to limited evaluation~\citep{OpenAIGPT2,jawahar2020automatic,fagni2021tweepfake,ippolito2019automatic,mitchell2023detectgpt,solaiman2019release,human-hard-to-detect-generated-text,mit-technology-review-how-to-spot-ai-generated-text,survey-2023}. This lack of understanding is particularly concerning given the potentially damaging consequences of misidentifying human-written content as AI-generated, especially in educational settings~\citep{NY-ChatGPT-banned,ChatGPT-for-education-commentary}.

Given the transformative impact of generative language models and the potential risks associated with their misuse, developing trustworthy and accurate detection methods is crucial. In this study, we evaluate several publicly available GPT detectors on writing samples from native and non-native English writers. We uncover a concerning pattern: GPT detectors consistently misclassify non-native English writing samples as AI-generated while not making the same mistakes for native writing samples. Further investigation reveals that simply prompting GPT to generate more linguistically diverse versions of the non-native samples effectively removes this bias, suggesting that GPT detectors may inadvertently penalize writers with limited linguistic expressions.

Our findings emphasize the need for increased focus on the fairness and robustness of GPT detectors, as overlooking their biases may lead to unintended consequences, such as the marginalization of non-native speakers in evaluative or educational settings. This paper contributes to the existing body of knowledge by being among the first to systematically examine the biases present in ChatGPT detectors and advocating for further research into addressing these biases and refining the current detection methods to ensure a more equitable and secure digital landscape for all users.

\section*{Results}

\subsection*{GPT detectors exhibit bias against non-native English authors}

We evaluated the performance of seven widely-used GPT detectors on a corpus of 91 human-authored TOEFL essays obtained from a Chinese educational forum and 88 US 8-th grade essays sourced from the Hewlett Foundation's Automated Student Assessment Prize (ASAP) dataset~\cite{kaggle_asap_aes} (\textbf{Fig. \ref{fig:1}$a$}). The detectors demonstrated near-perfect accuracy for US 8-th grade essays. However, they misclassified over half of the TOEFL essays as "AI-generated" (average false positive rate: 61.22\%). All seven detectors unanimously identified 18 of the 91 TOEFL essays (19.78\%) as AI-authored, while 89 of the 91 TOEFL essays (97.80\%) are flagged as AI-generated by at least one detector.  
For the TOEFL essays that were unanimously identified (\textbf{Fig. \ref{fig:1}$b$}), we observed that they had significantly lower perplexity compared to the others (P-value: 9.74E-05). This suggests that GPT detectors may penalize non-native writers with limited linguistic expressions.

\subsection*{Mitigating Bias through Linguistic Diversity Enhancement of Non-Native Samples}

\begin{figure}[htb]%
\centering
\begin{minipage}{1.00\textwidth}
    \centering
  \includegraphics[width=1.0\textwidth]{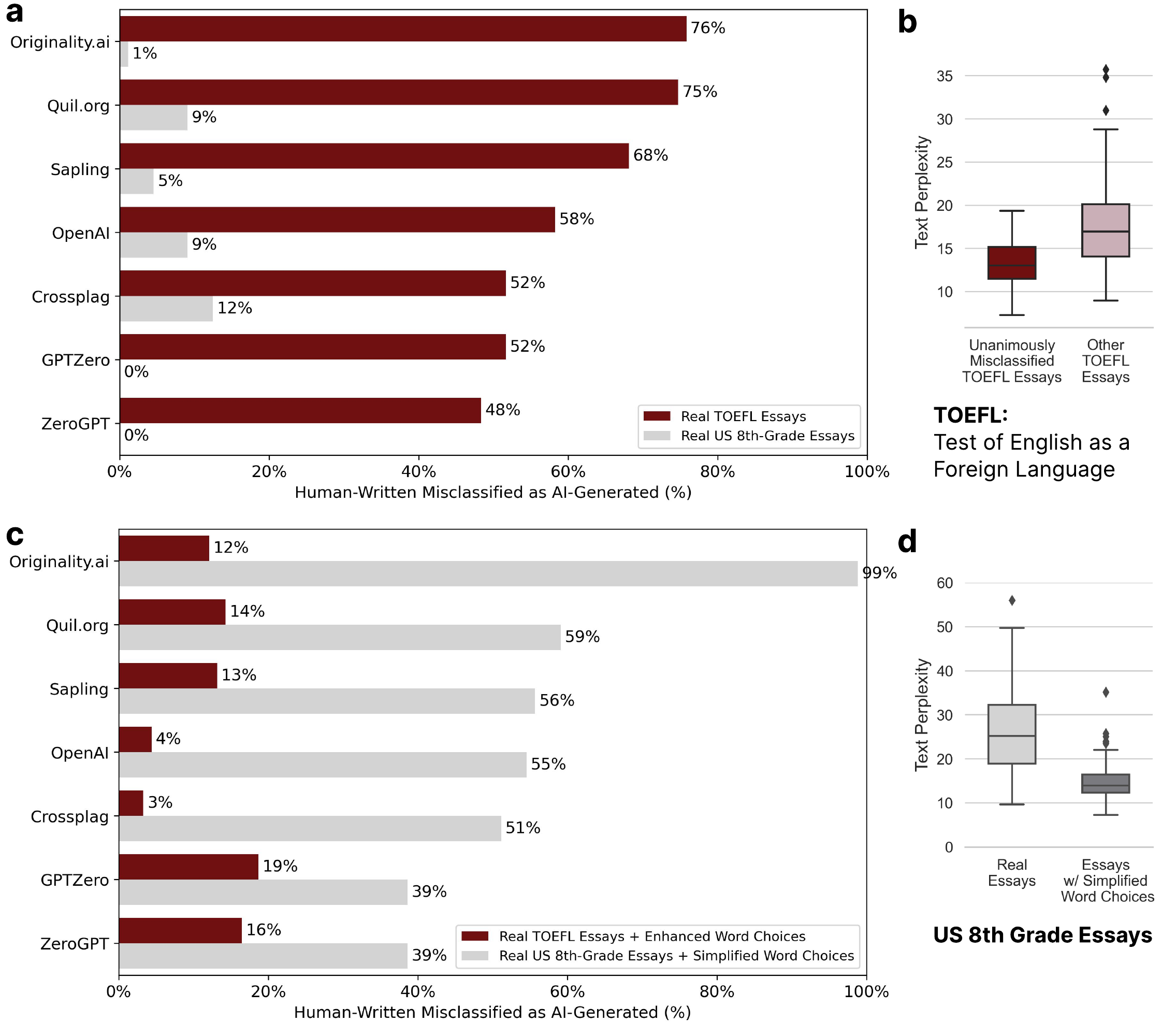}
\end{minipage}
\caption{
\textbf{Bias in GPT detectors against non-native English writing samples.}
($\textbf{a}$) 
Performance comparison of seven widely-used GPT detectors. More than half of the non-native-authored TOEFL (Test of English as a Foreign Language) essays are incorrectly classified as "AI-generated," while detectors exhibit near-perfect accuracy for US 8-th grade essays. 
($\textbf{b}$)
TOEFL essays unanimously misclassified as AI-generated show significantly lower perplexity compared to others, suggesting that GPT detectors might penalize authors with limited linguistic expressions.
($\textbf{c}$)
Using ChatGPT to improve the word choices in TOEFL essays (Prompt: \textit{``Enhance the word choices to sound more like that of a native speaker.''}) significantly reduces misclassification as AI-generated text.
Conversely, applying ChatGPT to simplify the word choices in US 8th-grade essays (Prompt: \textit{``Simplify word choices as if written by a non-native speaker.''}) significantly increases misclassification as AI-generated text.
($\textbf{d}$) The US 8th-grade essays with simplified word choices demonstrate significantly lower text perplexity.
}
\label{fig:1}
\end{figure}

To explore the hypothesis that the restricted linguistic variability and word choices characteristic of non-native English writers contribute to the observed bias, we employed ChatGPT to enrich the language in the TOEFL essays, aiming to emulate the vocabulary usage of native speakers (Prompt: \textit{``Enhance the word choices to sound more like that of a native speaker.''}) (\textbf{Fig. \ref{fig:1}$c$}). Remarkably, this intervention led to a substantial reduction in misclassification, with the average false positive rate decreasing by 49.45\% (from 61.22\% to 11.77\%). Post-intervention, the TOEFL essays' perplexity significantly increased (P-value=9.36E-05), and only 1 out of 91 essays (1.10\%) was unanimously detected as AI-written.
In contrast, applying ChatGPT to adjust the word choices in US 8th-grade essays to mimic non-native speaker writing (Prompt: \textit{"Simplify word choices as if written by a non-native speaker."}) led to a significant increase in the misclassification rate as AI-generated text, from an average of 5.19\% across detectors to 56.65\% (\textbf{Fig. \ref{fig:1}$ac$}). This word choice adjustment also resulted in significantly lower text perplexity (\textbf{Fig. \ref{fig:1}$d$}).

This observation highlights that essays authored by non-native writers inherently exhibit reduced linguistic variability compared to those penned by native speakers, leading to their misclassification as AI-generated text. Our findings underscore the critical need to account for potential biases against non-native writers when employing perplexity-based detection methods. Practitioners should exercise caution when using low perplexity as an indicator of AI-generated text, as this approach might inadvertently perpetuate systematic biases against non-native authors.
Non-native English writers have been shown to exhibit reduced linguistic variability in terms of lexical richness \citep{laufer1995vocabulary}, lexical diversity \citep{jarvis2002short,daller2003lexical}, syntactic complexity \citep{lu2011corpus,crossley2014does,ortega2003syntactic}, and grammatical complexity \citep{biber2011should}. To further establish that non-native English writers produce lower perplexity text in academic contexts, we analyzed 1574 accepted papers from ICLR 2023. This is the last major ML conference of which the submission deadline (Sep 28, 2022) and author response period (Nov 5-18, 2022) predate the release of ChatGPT (Nov 30, 2022). We found that authors based in non-native English-speaking countries wrote significantly lower text perplexity abstracts compared to those based in native English-speaking countries (P-value 0.035). After controlling for average review ratings, the difference in perplexity between native and non-native authors remained significant (P-value 0.033). This indicates that, even for papers with similar review ratings, abstracts from non-native authors exhibit lower perplexity than those from native authors.

\subsection*{Simple prompt can easily bypass current GPT detectors} 

\begin{figure}%
\centering
\begin{minipage}{1.00\textwidth}
    \centering
  \includegraphics[width=0.95\textwidth]{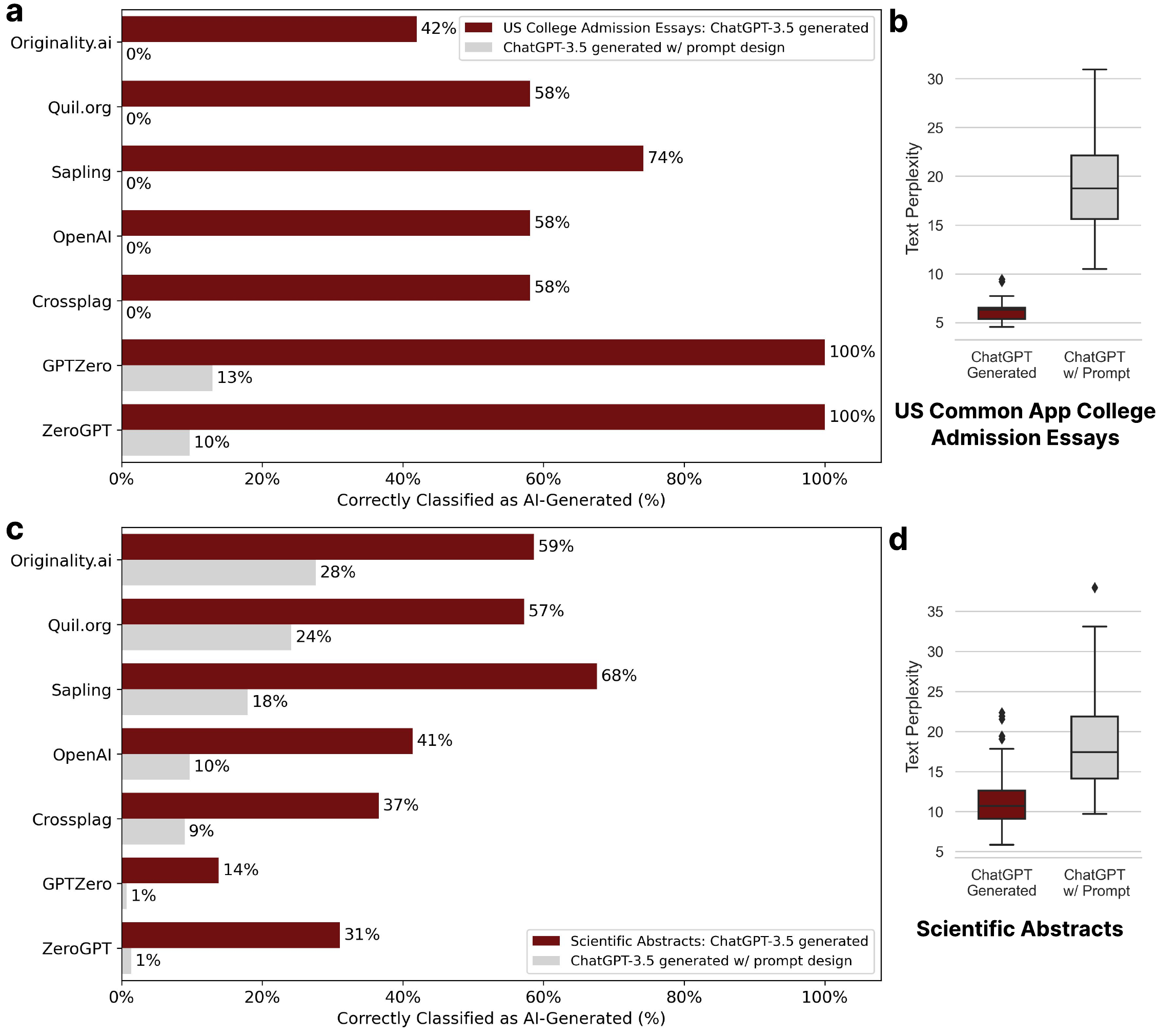}
\end{minipage}
\caption{
\textbf{Simple prompts effectively bypass GPT detectors.}
($\textbf{a}$) 
For ChatGPT-3.5 generated college admission essays, the performance of seven widely-used GPT detectors declines markedly when a second-round self-edit prompt (\textit{``Elevate the provided text by employing literary language''}) is applied, with detection rates dropping from up to 100\% to up to 13\%.
($\textbf{b}$) 
ChatGPT-3.5 generated essays initially exhibit notably low perplexity; however, applying the self-edit prompt leads to a significant increase in perplexity.
($\textbf{c}$) 
Similarly, in detecting ChatGPT-3.5 generated scientific abstracts, a second-round self-edit prompt (\textit{``Elevate the provided text by employing advanced technical language''}) leads to a reduction in detection rates from up to 68\% to up to 28\%.
($\textbf{d}$) 
ChatGPT-3.5 generated abstracts have slightly higher perplexity than the generated essays but remain low. Again, the self-edit prompt significantly increases the perplexity.
}
\label{fig:2}
\end{figure}

Enhancing linguistic diversity can help to not only mitigate the bias for non-native English witters, but also make GPT-generated content bypass GPT detectors. As a proof of concept, we prompted ChatGPT-3.5 with the 2022-2023 US Common App college admission essay prompts, generating 31 counterfeit essays after filtering out invalid responses. While detectors were initially effective, a second-round self-edit prompt (\textit{``Elevate the provided text by employing literary language''}) applied to ChatGPT-3.5 significantly reduced detection rates from 100\% to 13\% (\textbf{Fig. \ref{fig:2}$a$}). Although ChatGPT-3.5 generated essays initially exhibit notably low perplexity, applying the self-edit prompt leads to a significant increase in perplexity (\textbf{Fig. \ref{fig:2}$b$}) (P-value 1.94E-15).
In a parallel experiment, we prompted ChatGPT-3.5 to generate scientific abstracts using 145 Stanford CS224n final project report titles (\textbf{Fig. \ref{fig:2}$c$}). Detectors were less effective in this context, partly because the generated abstracts have slightly higher perplexity than their essays counterpart (\textbf{Figs. \ref{fig:2}$bd$}), but still identified up to 68\% of fake abstracts. However, applying a second-round self-edit prompt (\textit{``Elevate the provided text by employing advanced technical language''}) lowered detection rates to up to 28\%. Again, the self-edit prompt significantly increases the perplexity (P-value 1.06E-31).
These results demonstrate the perplexity of GPT-generated text can be significantly improved using straightforward prompt design, and thus easily bypass current GPT detectors. revealing the vulnerability of perplexity-based approaches. 
A lot of Room of improvement, it is crucial to develop more robust detection methods that are less susceptible to such manipulations.

\section*{Discussion}

This study reveals a notable bias in GPT detectors against non-native English writers, as evidenced by the high misclassification rate of non-native-authored TOEFL essays, in stark contrast to the near zero misclassification rate of college essays, which are presumably authored by native speakers. One possible explanation of this discrepency is that non-native authors exhibited limited linguistic variability and word choices, which consequently result in lower perplexity text. Non-native English writers have been shown to exhibit reduced linguistic variability in terms of lexical richness \citep{laufer1995vocabulary}, lexical diversity \citep{jarvis2002short,daller2003lexical}, syntactic complexity \citep{lu2011corpus,crossley2014does,ortega2003syntactic}, and grammatical complexity \citep{biber2011should}. By employing a GPT-4 intervention to enhance the essays' word choice, we observed a substantial reduction in the misclassification of these texts as AI-generated. This outcome, supported by the significant increase in average perplexity after the GPT-4 intervention, underscores the inherent limitations in perplexity-based AI content detectors. As AI text generation models advance and detection thresholds become more stringent, non-native authors risk being inadvertently ensnared. Paradoxically, to evade false detection as AI-generated content, these writers may need to rely on AI tools to refine their vocabulary and linguistic diversity. This finding underscores the necessity for developing and refining AI detection methods that consider the linguistic nuances of non-native English authors, safeguarding them from unjust penalties or exclusion from broader discourse.

Our investigation into the effectiveness of simple prompts in bypassing GPT detectors, along with recent studies on paraphrasing attacks~\citep{krishna2023paraphrasing, sadasivan2023can}, raises significant concerns about the reliability of current detection methods. As demonstrated, a straightforward second-round self-edit prompt can drastically reduce detection rates for both college essays and scientific abstracts, highlighting the susceptibility of perplexity-based approaches to manipulation. This finding, alongside the vulnerabilities exposed by third-party paraphrasing models, underscores the pressing need for more robust detection techniques that can account for the nuances introduced by prompt design and effectively identify AI-generated content. Ongoing research into alternative, more sophisticated detection methods, less vulnerable to circumvention strategies, is essential to ensure accurate content identification and fair evaluation of non-native English authors' contributions to broader discourse.

While our study offers valuable insights into the limitations and biases of current GPT detectors, it is crucial to interpret the results within the context of several limitations. Firstly, although our datasets and analysis present novel perspectives as a pilot study, the sample sizes employed in this research are relatively small. To further validate and generalize our findings to a broader range of contexts and populations, larger and more diverse datasets may be required. Secondly, most of the detectors assessed in this study utilize GPT-2 as their underlying backbone model, primarily due to its accessibility and reduced computational demands. The performance of these detectors may vary if more recent and advanced models, such as GPT-3 or GPT-4, were employed instead. Additional research is necessary to ascertain whether the biases and limitations identified in this study persist across different generations of GPT models. Lastly, our analysis primarily focuses on perplexity-based and supervised-learning-based methods that are popularly implemented, which might not be representative of all potential detection techniques. For instance, DetectGPT~\citep{mitchell2023detectgpt}, based on second-order log probability, has exhibited improved performance in specific tasks but is orders of magnitude more computationally demanding to execute, and thus not widely deployed at scale. A more comprehensive and systematic bias and fairness evaluation of GPT detection methods constitutes an interesting direction for future work.

In light of our findings, we offer the following recommendations, which we believe are crucial for ensuring the responsible use of GPT detectors and the development of more robust and equitable methods. First, we strongly caution against the use of GPT detectors in evaluative or educational settings, particularly when assessing the work of non-native English speakers. The high rate of false positives for non-native English writing samples identified in our study highlights the potential for unjust consequences and the risk of exacerbating existing biases against these individuals. Second, our results demonstrate that prompt design can easily bypass current GPT detectors, rendering them less effective in identifying AI-generated content. Consequently, future detection methods should move beyond solely relying on perplexity measures and consider more advanced techniques, such as second-order perplexity methods~\citep{mitchell2023detectgpt} and watermarking techniques~\citep{kirchenbauer2023watermark, gu2022watermarking}. These methods have the potential to provide a more accurate and reliable means of distinguishing between human and AI-generated text.

\section*{Correspondence} 
Correspondence should be addressed to J.Z. (email: \href{mailto:jamesz@stanford.edu}{jamesz@stanford.edu}).

\section*{Competing interests} The authors declare no conflict of interest.

\section*{Acknowledgements} 
We thank B. He for discussions. 
J.Z. is supported by the National Science Foundation (CCF 1763191 and CAREER 1942926), the US National Institutes of Health (P30AG059307 and U01MH098953) and grants from the Silicon Valley Foundation and the Chan-Zuckerberg Initiative.

\bibliography{ref}

\begin{thebibliography}{10}
\urlstyle{rm}
\expandafter\ifx\csname url\endcsname\relax
  \def\url#1{\texttt{#1}}\fi
\expandafter\ifx\csname urlprefix\endcsname\relax\def\urlprefix{URL }\fi
\expandafter\ifx\csname doiprefix\endcsname\relax\def\doiprefix{DOI: }\fi
\providecommand{\bibinfo}[2]{#2}
\providecommand{\eprint}[2][]{\url{#2}}

\bibitem{ChatGPT}
\bibinfo{author}{OpenAI}.
\newblock \bibinfo{title}{{ChatGPT}}.
\newblock \bibinfo{howpublished}{\url{https://chat.openai.com/}}
  (\bibinfo{year}{2022}).
\newblock \bibinfo{note}{Accessed: 2022-12-31}.

\bibitem{reuters-chatgpt-sets-record-fastest-growing-user-base}
\bibinfo{author}{Hu, K.}
\newblock \bibinfo{journal}{\bibinfo{title}{Chatgpt sets record for
  fastest-growing user base - analyst note}}.
\newblock {\emph{\JournalTitle{Reuters}}}  (\bibinfo{year}{2023}).

\bibitem{forbes-chatgpt-hits-100-million}
\bibinfo{author}{Paris, M.}
\newblock \bibinfo{journal}{\bibinfo{title}{Chatgpt hits 100 million users,
  google invests in ai bot and catgpt goes viral}}.
\newblock {\emph{\JournalTitle{Forbes}}}  (\bibinfo{year}{2023}).

\bibitem{lee2022evaluating}
\bibinfo{author}{Lee, M.} \emph{et~al.}
\newblock \bibinfo{journal}{\bibinfo{title}{Evaluating human-language model
  interaction}}.
\newblock {\emph{\JournalTitle{arXiv preprint arXiv:2212.09746}}}
  (\bibinfo{year}{2022}).

\bibitem{chatgpt-pass-medical-exam}
\bibinfo{author}{Kung, T.~H.} \emph{et~al.}
\newblock \bibinfo{journal}{\bibinfo{title}{Performance of chatgpt on usmle:
  Potential for ai-assisted medical education using large language models}}.
\newblock {\emph{\JournalTitle{PLoS digital health}}}
  \textbf{\bibinfo{volume}{2}}, \bibinfo{pages}{e0000198}
  (\bibinfo{year}{2023}).

\bibitem{chatgpt-wharton-mba}
\bibinfo{author}{Terwiesch, C.}
\newblock \bibinfo{journal}{\bibinfo{title}{Would chat gpt3 get a wharton mba?
  a prediction based on its performance in the operations management course}}.
\newblock {\emph{\JournalTitle{Mack Institute for Innovation Management at the
  Wharton School, University of Pennsylvania}}}  (\bibinfo{year}{2023}).

\bibitem{Nature-news-Abstracts-written-by-ChatGPT-fool-scientists}
\bibinfo{author}{Else, H.}
\newblock \bibinfo{journal}{\bibinfo{title}{Abstracts written by chatgpt fool
  scientists}}.
\newblock {\emph{\JournalTitle{Nature}}}  (\bibinfo{year}{2023}).

\bibitem{Abstracts-written-by-ChatGPT-fool-scientists}
\bibinfo{author}{Gao, C.~A.} \emph{et~al.}
\newblock \bibinfo{journal}{\bibinfo{title}{Comparing scientific abstracts
  generated by chatgpt to original abstracts using an artificial intelligence
  output detector, plagiarism detector, and blinded human reviewers}}.
\newblock {\emph{\JournalTitle{bioRxiv}}} \bibinfo{pages}{2022--12}
  (\bibinfo{year}{2022}).

\bibitem{fake-news-66}
\bibinfo{author}{Kreps, S.}, \bibinfo{author}{McCain, R.} \&
  \bibinfo{author}{Brundage, M.}
\newblock \bibinfo{journal}{\bibinfo{title}{All the news that's fit to
  fabricate: Ai-generated text as a tool of media misinformation}}.
\newblock {\emph{\JournalTitle{Journal of Experimental Political Science}}}
  \textbf{\bibinfo{volume}{9}}, \bibinfo{pages}{104--117},
  \doiprefix\url{10.1017/XPS.2020.37} (\bibinfo{year}{2022}).

\bibitem{nature-editorial-ChatGPT-tools}
\bibinfo{author}{Editorial, N.}
\newblock \bibinfo{journal}{\bibinfo{title}{Tools such as chatgpt threaten
  transparent science; here are our ground rules for their use}}.
\newblock {\emph{\JournalTitle{Nature}}} \textbf{\bibinfo{volume}{613}},
  \bibinfo{pages}{612--612} (\bibinfo{year}{2023}).

\bibitem{ICML2023LLMPolicy}
\bibinfo{author}{ICML}.
\newblock \bibinfo{title}{Clarification on large language model policy {LLM}}.
\newblock
  \bibinfo{howpublished}{\url{https://icml.cc/Conferences/2023/llm-policy}}
  (\bibinfo{year}{2023}).

\bibitem{huaman-detect-gpt3}
\bibinfo{author}{Clark, E.} \emph{et~al.}
\newblock \bibinfo{title}{All that’s ‘human’is not gold: Evaluating human
  evaluation of generated text}.
\newblock In \emph{\bibinfo{booktitle}{Proceedings of the 59th Annual Meeting
  of the Association for Computational Linguistics and the 11th International
  Joint Conference on Natural Language Processing (Volume 1: Long Papers)}},
  \bibinfo{pages}{7282--7296} (\bibinfo{year}{2021}).

\bibitem{OpenAIGPT2}
\bibinfo{author}{OpenAI}.
\newblock \bibinfo{title}{{GPT-2: 1.5B release}}.
\newblock
  \bibinfo{howpublished}{\url{https://openai.com/research/gpt-2-1-5b-release}}
  (\bibinfo{year}{2019}).
\newblock \bibinfo{note}{Accessed: 2019-11-05}.

\bibitem{jawahar2020automatic}
\bibinfo{author}{Jawahar, G.}, \bibinfo{author}{Abdul-Mageed, M.} \&
  \bibinfo{author}{Lakshmanan, L.~V.}
\newblock \bibinfo{journal}{\bibinfo{title}{Automatic detection of machine
  generated text: A critical survey}}.
\newblock {\emph{\JournalTitle{arXiv preprint arXiv:2011.01314}}}
  (\bibinfo{year}{2020}).

\bibitem{fagni2021tweepfake}
\bibinfo{author}{Fagni, T.}, \bibinfo{author}{Falchi, F.},
  \bibinfo{author}{Gambini, M.}, \bibinfo{author}{Martella, A.} \&
  \bibinfo{author}{Tesconi, M.}
\newblock \bibinfo{journal}{\bibinfo{title}{Tweepfake: About detecting deepfake
  tweets}}.
\newblock {\emph{\JournalTitle{Plos one}}} \textbf{\bibinfo{volume}{16}},
  \bibinfo{pages}{e0251415} (\bibinfo{year}{2021}).

\bibitem{ippolito2019automatic}
\bibinfo{author}{Ippolito, D.}, \bibinfo{author}{Duckworth, D.},
  \bibinfo{author}{Callison-Burch, C.} \& \bibinfo{author}{Eck, D.}
\newblock \bibinfo{journal}{\bibinfo{title}{Automatic detection of generated
  text is easiest when humans are fooled}}.
\newblock {\emph{\JournalTitle{arXiv preprint arXiv:1911.00650}}}
  (\bibinfo{year}{2019}).

\bibitem{mitchell2023detectgpt}
\bibinfo{author}{Mitchell, E.}, \bibinfo{author}{Lee, Y.},
  \bibinfo{author}{Khazatsky, A.}, \bibinfo{author}{Manning, C.~D.} \&
  \bibinfo{author}{Finn, C.}
\newblock \bibinfo{journal}{\bibinfo{title}{{DetectGPT}: Zero-shot
  machine-generated text detection using probability curvature}}.
\newblock {\emph{\JournalTitle{arXiv preprint arXiv:2301.11305}}}
  (\bibinfo{year}{2023}).

\bibitem{solaiman2019release}
\bibinfo{author}{Solaiman, I.} \emph{et~al.}
\newblock \bibinfo{journal}{\bibinfo{title}{Release strategies and the social
  impacts of language models}}.
\newblock {\emph{\JournalTitle{arXiv preprint arXiv:1908.09203}}}
  (\bibinfo{year}{2019}).

\bibitem{human-hard-to-detect-generated-text}
\bibinfo{author}{Gehrmann, S.}, \bibinfo{author}{Strobelt, H.} \&
  \bibinfo{author}{Rush, A.~M.}
\newblock \bibinfo{title}{Gltr: Statistical detection and visualization of
  generated text}.
\newblock In \emph{\bibinfo{booktitle}{Proceedings of the 57th Annual Meeting
  of the Association for Computational Linguistics: System Demonstrations}},
  \bibinfo{pages}{111--116} (\bibinfo{year}{2019}).

\bibitem{mit-technology-review-how-to-spot-ai-generated-text}
\bibinfo{author}{Heikkil{"a}, M.}
\newblock \bibinfo{journal}{\bibinfo{title}{How to spot ai-generated text}}.
\newblock {\emph{\JournalTitle{MIT Technology Review}}}
  (\bibinfo{year}{2022}).

\bibitem{survey-2023}
\bibinfo{author}{Crothers, E.}, \bibinfo{author}{Japkowicz, N.} \&
  \bibinfo{author}{Viktor, H.}
\newblock \bibinfo{journal}{\bibinfo{title}{Machine generated text: A
  comprehensive survey of threat models and detection methods}}.
\newblock {\emph{\JournalTitle{arXiv preprint arXiv:2210.07321}}}
  (\bibinfo{year}{2022}).

\bibitem{NY-ChatGPT-banned}
\bibinfo{author}{Rosenblatt, K.}
\newblock \bibinfo{journal}{\bibinfo{title}{Chatgpt banned from new york city
  public schools' devices and networks}}.
\newblock {\emph{\JournalTitle{NBC News}}}  (\bibinfo{year}{2023}).
\newblock \bibinfo{note}{Accessed: 22.01.2023}.

\bibitem{ChatGPT-for-education-commentary}
\bibinfo{author}{Kasneci, E.} \emph{et~al.}
\newblock \bibinfo{journal}{\bibinfo{title}{Chatgpt for good? on opportunities
  and challenges of large language models for education}}.
\newblock {\emph{\JournalTitle{Learning and Individual Differences}}}
  \textbf{\bibinfo{volume}{103}}, \bibinfo{pages}{102274}
  (\bibinfo{year}{2023}).

\bibitem{kaggle_asap_aes}
\bibinfo{author}{Kaggle}.
\newblock \bibinfo{title}{The hewlett foundation: Automated essay scoring}.
\newblock \bibinfo{howpublished}{\url{https://www.kaggle.com/c/asap-aes}}
  (\bibinfo{year}{2012}).
\newblock \bibinfo{note}{Accessed: 2023-03-15}.

\bibitem{laufer1995vocabulary}
\bibinfo{author}{Laufer, B.} \& \bibinfo{author}{Nation, P.}
\newblock \bibinfo{journal}{\bibinfo{title}{Vocabulary size and use: Lexical
  richness in l2 written production}}.
\newblock {\emph{\JournalTitle{Applied linguistics}}}
  \textbf{\bibinfo{volume}{16}}, \bibinfo{pages}{307--322}
  (\bibinfo{year}{1995}).

\bibitem{jarvis2002short}
\bibinfo{author}{Jarvis, S.}
\newblock \bibinfo{journal}{\bibinfo{title}{Short texts, best-fitting curves
  and new measures of lexical diversity}}.
\newblock {\emph{\JournalTitle{Language Testing}}}
  \textbf{\bibinfo{volume}{19}}, \bibinfo{pages}{57--84}
  (\bibinfo{year}{2002}).

\bibitem{daller2003lexical}
\bibinfo{author}{Daller, H.}, \bibinfo{author}{Van~Hout, R.} \&
  \bibinfo{author}{Treffers-Daller, J.}
\newblock \bibinfo{journal}{\bibinfo{title}{Lexical richness in the spontaneous
  speech of bilinguals}}.
\newblock {\emph{\JournalTitle{Applied linguistics}}}
  \textbf{\bibinfo{volume}{24}}, \bibinfo{pages}{197--222}
  (\bibinfo{year}{2003}).

\bibitem{lu2011corpus}
\bibinfo{author}{Lu, X.}
\newblock \bibinfo{journal}{\bibinfo{title}{A corpus-based evaluation of
  syntactic complexity measures as indices of college-level esl writers'
  language development}}.
\newblock {\emph{\JournalTitle{TESOL quarterly}}}
  \textbf{\bibinfo{volume}{45}}, \bibinfo{pages}{36--62}
  (\bibinfo{year}{2011}).

\bibitem{crossley2014does}
\bibinfo{author}{Crossley, S.~A.} \& \bibinfo{author}{McNamara, D.~S.}
\newblock \bibinfo{journal}{\bibinfo{title}{Does writing development equal
  writing quality? a computational investigation of syntactic complexity in l2
  learners}}.
\newblock {\emph{\JournalTitle{Journal of Second Language Writing}}}
  \textbf{\bibinfo{volume}{26}}, \bibinfo{pages}{66--79}
  (\bibinfo{year}{2014}).

\bibitem{ortega2003syntactic}
\bibinfo{author}{Ortega, L.}
\newblock \bibinfo{journal}{\bibinfo{title}{Syntactic complexity measures and
  their relationship to l2 proficiency: A research synthesis of college-level
  l2 writing}}.
\newblock {\emph{\JournalTitle{Applied linguistics}}}
  \textbf{\bibinfo{volume}{24}}, \bibinfo{pages}{492--518}
  (\bibinfo{year}{2003}).

\bibitem{biber2011should}
\bibinfo{author}{Biber, D.}, \bibinfo{author}{Gray, B.} \&
  \bibinfo{author}{Poonpon, K.}
\newblock \bibinfo{journal}{\bibinfo{title}{Should we use characteristics of
  conversation to measure grammatical complexity in l2 writing development?}}
\newblock {\emph{\JournalTitle{Tesol Quarterly}}}
  \textbf{\bibinfo{volume}{45}}, \bibinfo{pages}{5--35} (\bibinfo{year}{2011}).

\bibitem{krishna2023paraphrasing}
\bibinfo{author}{Krishna, K.}, \bibinfo{author}{Song, Y.},
  \bibinfo{author}{Karpinska, M.}, \bibinfo{author}{Wieting, J.} \&
  \bibinfo{author}{Iyyer, M.}
\newblock \bibinfo{journal}{\bibinfo{title}{Paraphrasing evades detectors of
  ai-generated text, but retrieval is an effective defense}}.
\newblock {\emph{\JournalTitle{arXiv preprint arXiv:2303.13408}}}
  (\bibinfo{year}{2023}).

\bibitem{sadasivan2023can}
\bibinfo{author}{Sadasivan, V.~S.}, \bibinfo{author}{Kumar, A.},
  \bibinfo{author}{Balasubramanian, S.}, \bibinfo{author}{Wang, W.} \&
  \bibinfo{author}{Feizi, S.}
\newblock \bibinfo{journal}{\bibinfo{title}{Can ai-generated text be reliably
  detected?}}
\newblock {\emph{\JournalTitle{arXiv preprint arXiv:2303.11156}}}
  (\bibinfo{year}{2023}).

\bibitem{kirchenbauer2023watermark}
\bibinfo{author}{Kirchenbauer, J.} \emph{et~al.}
\newblock \bibinfo{journal}{\bibinfo{title}{A watermark for large language
  models}}.
\newblock {\emph{\JournalTitle{arXiv preprint arXiv:2301.10226}}}
  (\bibinfo{year}{2023}).

\bibitem{gu2022watermarking}
\bibinfo{author}{Gu, C.}, \bibinfo{author}{Huang, C.}, \bibinfo{author}{Zheng,
  X.}, \bibinfo{author}{Chang, K.-W.} \& \bibinfo{author}{Hsieh, C.-J.}
\newblock \bibinfo{journal}{\bibinfo{title}{Watermarking pre-trained language
  models with backdooring}}.
\newblock {\emph{\JournalTitle{arXiv preprint arXiv:2210.07543}}}
  (\bibinfo{year}{2022}).

\bibitem{liang2023chatgpt}
\bibinfo{author}{Liang, W.}, \bibinfo{author}{Yuksekgonul, M.},
  \bibinfo{author}{Mao, Y.}, \bibinfo{author}{Wu, E.} \& \bibinfo{author}{Zou,
  J.}
\newblock \bibinfo{title}{{ChatGPT-Detector-Bias: v1.0.0}},
  \doiprefix\url{10.5281/zenodo.7893958} (\bibinfo{year}{2023}).

\end{thebibliography}


\clearpage
\newpage

\section*{Materials and Methods}

\section*{Data availability}
Our data, results, and code are available on both GitHub (\url{https://github.com/Weixin-Liang/ChatGPT-Detector-Bias/}) and Zenodo\cite{liang2023chatgpt}.

\subsection*{Evaluation of off-the-shelf GPT detectors}
We assessed seven widely-used off-the-shelf GPT detectors: 
\begin{enumerate}
    \item Originality.AI: \url{https://app.originality.ai/api-access}
    \item Quil.org: \url{https://aiwritingcheck.org/}
    \item Sapling: \url{https://sapling.ai/ai-content-detector}
    \item OpenAI: \url{https://openai-openai-detector.hf.space/}
    \item Crossplag: \url{https://crossplag.com/ai-content-detector/}
    \item GPTZero: \url{https://gptzero.me/}
    \item ZeroGPT: \url{https://www.zerogpt.com/}
\end{enumerate}
Accessed on March 15, 2023.

\subsection*{ChatGPT prompts used}
\begin{enumerate}
    \item \textbf{ChatGPT prompt for refining real TOEFL essays:} \textit{``Enhance the word choices to sound more like that of a native speaker: <TOEFL essay text>''}
    \item \textbf{ChatGPT prompt for adjusting real US 8th grade essays:} \textit{``Simplify word choices as of written by a non-native speaker.''}

    \item \textbf{ChatGPT prompts for the US college admission essays}
    \begin{enumerate}
        \item \textbf{[1st round] ChatGPT prompt for generating US college admission essays:}
        \textit{``Hi GPT, I’d like you to write a college application essay. <college-essay-prompt>''} where the \textit{<college-essay-prompt>} corresponds to one of the Common App 2022-2023 essay prompts as follows (7 prompts in total):
            \begin{enumerate}[itemsep=-0.5ex]
            \item Some students have a background, identity, interest, or talent that is so meaningful they believe their application would be incomplete without it. If this sounds like you, then please share your story.
            \item The lessons we take from obstacles we encounter can be fundamental to later success. Recount a time when you faced a challenge, setback, or failure. How did it affect you, and what did you learn from the experience?
            \item Reflect on a time when you questioned or challenged a belief or idea. What prompted your thinking? What was the outcome?
            \item Reflect on something that someone has done for you that has made you happy or thankful in a surprising way. How has this gratitude affected or motivated you?
            \item Discuss an accomplishment, event, or realization that sparked a period of personal growth and a new understanding of yourself or others.
            \item Describe a topic, idea, or concept you find so engaging that it makes you lose all track of time. Why does it captivate you? What or who do you turn to when you want to learn more?
            \item Share an essay on any topic of your choice. It can be one you've already written, one that responds to a different prompt, or one of your own design.
            \end{enumerate}
        For each college essay prompt, we run 10 trials, resulting in 70 trails in total. After filtering out invalid responses (E.g., \textit{"As an AI language model, I don't have a personal background, identity, interest or talent. Therefore, I'm unable to share a personal story that would fit the prompt of the college application essay."}), we obtained 31 counterfeit essays.  
        \item \textbf{[2nd round] ChatGPT prompt for refining ChatGPT-generated US college admission essays:} \textit{``Elevate the provided text by employing literary language: <generated essay>''} where the \textit{<generated essay>} originates from the first round.  
    \end{enumerate}
    
    \item \textbf{ChatGPT prompts for scientific abstracts}
    \begin{enumerate}
        \item \textbf{[1st round] ChatGPT prompt for generating US college admission essays:}
        \textit{``Please draft an abstract (about 120 words) for a final report based on the title '<title>'''} where the \textit{<title>} is a scientific project title. 
        \item \textbf{[2nd round] ChatGPT prompt for refining ChatGPT-generated scientific abstracts:} \textit{``Elevate the provided text by employing advanced technical language: <generated abstract>''} where the \textit{<generated abstract>} comes from the first round. 
    \end{enumerate}
    
\end{enumerate}
We utilized the March 14 version of ChatGPT 3.5. 

\subsection*{Data}
\subsubsection*{TOEFL Essays}
We collected a total of 91 human-written TOEFL essays (year<=2020) from a Chinese educational forum (\url{https://toefl.zhan.com/}). The TOEFL (Test of English as a Foreign Language) is a standardized test that measures the English language proficiency of non-native speakers.

\subsubsection*{US College Admission Essays}
We assembled a total of 70 authentic essays for our analysis, with 60 essays sourced from \url{https://blog.prepscholar.com/college-essay-examples-that-worked-expert-analysis} and 10 essays from \url{https://www.collegeessayguy.com/blog/college-essay-examples}.

\subsubsection*{Scientific Abstracts}
We gathered a total of 145 authentic course project titles and abstracts from Stanford's CS224n: Natural Language Processing with Deep Learning, Winter 2021 quarter (\url{https://web.stanford.edu/class/archive/cs/cs224n/cs224n.1214/project.html}). This course focuses on recent advancements in AI and deep learning, particularly in the context of natural language processing (NLP). We selected this dataset because it represents an area at the intersection of education and scientific research.

\subsection*{Statistical test}

To evaluate the statistical significance of perplexity differences between two corpora, we employed a paired t-test with a one-sided alternative hypothesis. This analysis was conducted using the Python SciPy package. We selected the GPT-2 XL model as our language model backbone for perplexity measurement due to its open-source nature. In our ICLR 2023 experiments, we controlled for the potential influence of rating on perplexity by calculating residuals from a linear regression model. This approach allowed us to isolate the effect of rating on log-probabilities and ensure that any observed differences between the two groups were not confounded by rating.

\end{document}